\documentclass[11pt,a4paper]{article}
\usepackage[hyperref]{naaclhlt2019}
\usepackage{times}
\usepackage{latexsym}
\usepackage{CJKutf8}  
\usepackage{pinyin}
\usepackage{graphicx}
\usepackage{amsmath}
\usepackage{url}

\aclfinalcopy 

\title{Emotion Action Detection and Emotion Inference: the Task and Dataset}

\author{Pengyuan Liu , Chengyu Du , Shuofeng Zhao\\
  Beijing Language and Culture University \\  
    {liupengyuan@pku.edu.cn, 2550611409@qq.com, ShuofengZhao@163.com}  }

\date{}
\begin{document}
\maketitle
\begin{abstract}
Many NLP works on emotion analysis only focus on simple emotion classification without exploring the potentials of putting emotion into “event context", and ignore the analysis of emotion-related events. One main reason is the lack of this kind of corpus. Here we present CEAC (Cause-Emotion-Action Corpus), which manually annotates not only emotion, but also cause events and action events. For example, \textit{“After $\langle {cause}\rangle$listening to what I said$\langle {/cause}\rangle$, the teacher was $\langle {emotion}\rangle$happy$\langle {/emotion}\rangle$ and then $\langle {action}\rangle$joked with me$\langle {/action}\rangle$."} And we propose two new tasks based on CEAC: emotion causality and emotion inference. The first task is to extract a triple (cause, emotion, action) as CEA relation. The second task is to infer the probable emotion(here tends to be “happy") given a tuple of cause and action events (\textit{“listening to what I said", “joked with me"}). We are currently releasing CEAC with 10,603 samples and 15,892 events, basic statistic analysis and  baseline on both emotion causality and emotion inference tasks. Baseline performance demonstrates that there is much room for both tasks to be improved.
\end{abstract}

\section{Introduction}
Understanding a text especially a narrative involving people’s emotions needs to analysis it from all aspects and it usually requires commonsense reasoning. For example, for a given text with emotion and corresponding event \textit{“After listening to what I said, the teacher was happy and then joked with me."}, to analysis the casuality, one can easily know \textit{“listening to what I said"} can be regarded as the cause which leads to the emotion \textit{happy}, \textit{“joked with me"} can be seen as the result or consequence (we name it as action) caused by the \textit{happy}. In fact, the triple (\textit{“listening to what I said"}, \textit{“happy”}, \textit{“joked with me"}) comprised a three-folds cause-effect chain (we name it as CEA: Cause-Emotion-Action relation) in which “happy" is the intermediate variable. On the other hand, to infer the emotion by given cause \textit{“listening to what I said"}, we hardly deduce that the teacher's emotion is happy or angry. But if we know the action \textit{“joked with me"}, which is caused by teacher's emotion, we can infer that teacher feel happy rather than angry. Actions as new emotion knowledge or common sense can help infer emotion.

\begin{figure}[t]
    \centering
    \includegraphics[width=3.0in]{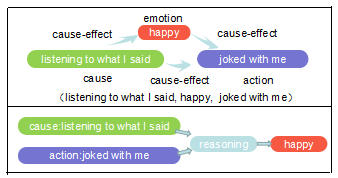}
    \caption{Emotion Causality and Emotion Inference}
\end{figure}

We present CEAC, a corpus which manually annotates not only emotion, but also emotion cause events and emotion action events. Based on CEAC, we introduce emotion causality and emotion inference task. The first task is to extract a triple of cause event, emotion and action event, and the second is to infer emotion given cause and action events(See Fig. 1). Our works are inspired by but quite different from previous researches. For instance, \cite{lee2010emotion} built a corpus, annotated emotion cause and proposed an emotion cause detection task; \cite{gui2016event} applied this method to Weibo, but he didn’t propose any new task;  \cite{xiyaochengargument} focused on current/original-subtweet-based emotion detection and annotated a multiple-user structure; \cite{deng2013benefactive} and \cite{ding2016acquiring} which introduced benefactive/malefactive events and defined affective events.

\indent Causal relation or causality are fundamental in many disciplines, including philosophy, psychology and linguistics. As one kind of event causality, emotion causality is also critical knowledge for many NLP applications, including machine reading and comprehension \cite{richardson2013mctest}, process extraction \cite{scaria2013learning}, and especially future event/scenario prediction \cite{radinsky2012learning}. Knowing the existence of an emotion is often insufficient to predict future events or determine the best reaction \cite{chen2010emotion}, whereas if the emotion cause and action is known to the corresponding emotion, prediction of future events or assessment of potential intent can be done more reliably. Furthermore, emotion inference and emotion causality are useful for a wide range of NLP applications that require anticipation of people’s emotional reaction and intents, especially when they are not explicitly mentioned. For example, an ideal dialogue system should react in empathetic ways by reasoning about the human user’s mental state based on the events the user has experienced, without the user explicitly stating how they are feeling \cite{rashkin2018event2mind}. Advertisement systems on social media should be able to reason about the emotional reactions of people after events such as mass shootings and remove ads for guns which might increase social distress  Goel and Isaac\footnote {https://www.nytimes.com/2016/01/30/technology/facebook-gun-sales-ban.html};\cite{rashkin2018event2mind}. Also, as one kind of pragmatic inference, emotion inference is a necessary step toward automatic narrative understanding and generation \cite{tomai2010using,ding2016acquiring,ding2018weakly}.

\indent Our contribution in this paper is threefold: 1) we define emotion action and put it into emotion causality so that cause, emotion and action comprise an integral cause-effect chain; 2) we define and investigate emotion causality and emotion inference tasks to bridge the gap between the study of emotion cause, affective events and commonsense inference; 
3) we manually label a large-scale corpus containing not only emotion, but also emotion cause events and action events.
 
\section{Construction of CEAC}

\subsection{Term Definition}
\textbf{Emotion }is the interrelated, synchronized changes in the states of all or most of the organismic subsystems in response to the evaluation of an external or internal stimulus event as relevant to major concerns of the organism \cite{scherer2005emotions}.

\noindent \textbf{Experiencer} is the person or sentient entity who has a particular emotional state. \cite{fillmore2003framenet}.

\noindent \textbf{Emotion cause} refers to the event that evokes the emotional response in the Experiencer. Our definition is similar to \cite{lee2010emotion} where she called cause event and she regarded it refers to the immediate cause of the emotion, which can be the actual trigger event or the perception of the trigger event. We think her definition ignores the experiencer. Emotion cause is similar to the emotion-provoking event \cite{tokuhisa2008emotion} but he did not make clear definition. It is also like the definition of emotion stimulus \cite{fillmore2003framenet} but we limit the cause to the event.

\noindent \textbf{Emotion action} refers to the event carried out by Experiencer that reflects his or her emotion state or emotion change. We use emotion action rather than emotion expression \cite{charles1872expression} because the latter focused on the facial expression, behavioral response, and physical responses  of the experiencer whereas we care more about the event.
\subsection{Data Collection}

\subsubsection{Taxonomy of emotion.} We adopt Ekman’s emotion classification \cite{ekman1992argument}, which identifies six primary emotions, namely happiness, sadness, fear, anger, disgust and surprise. This list is agreed upon by most previous works including Chinese emotion analysis, so the use of this list contributes to resource sharing.

\subsubsection{Emotion keywords.} We plan to construct CEAC in two stages. The first stage is to build about 10000 instances with representative emotion keywords which is the work of this paper. The second stage in future is to build about 40,000 instances with all the words from the existing Chinese emotional dictionary. Here we introduce the selection steps of representative emotion keywords on first stage. 

In Scherer's components processing model of emotion, five crucial elements of emotion are said to exist, of which feeling is the subjective experience of emotional state once it has occurred \cite{scherer2005emotions}. So people can feel emotions and the emotion keywords within the format “feel emotion”\footnote{Unlike English, Most frequently used Chinese emotion words are suitable for the format.} are more representative in the text. So the steps for choosing emotion keywords are as follows:
\begin{enumerate}
\item Find the intersection emotion keyword set (The single Chinese character word are excluded to avoid strong sense ambiguity) among the three Chinese emotion dictionary: the emotion list of Hownet\footnote{http://www.keenage.com}, the emotional\_word\_ontology\footnote{http://ir.dlut.edu.cn/EmotionOntologyDownload} and NTUSD\footnote{http://academiasinicanlplab.github.io/}.
\item To all the words in that word set, count the 2-gram “\begin{CJK}{UTF8}{gkai}
感到\end{CJK}”/“feel”+“\begin{CJK}{UTF8}{gkai}情感词\end{CJK}”/“emotion”  such as “\begin{CJK}{UTF8}{gkai}感到高兴\end{CJK}”/“feel happy”.
\item Choose the top-5 frequency 2-grams for each emotion category, delete the word “\begin{CJK}{UTF8}{gkai}感到\end{CJK}”/“feel”, then get the emotion keywords.
\end{enumerate}
Finally, 30 emotion keywords are selected as showed in table 1 below, followed by its English translation.

\begin{table*}[ht]
\centering  
\begin{tabular}{ll}  
\hline
\textbf{Emotion category} & \textbf{Emotion keywords.} \\ \hline
Happiness & \begin{CJK}{UTF8}{gkai}快乐、高兴、欢乐、开心、愉快\end{CJK} \\ 
 & Happy, pleased, joyful, cheerful, merry \\
 Sadness & \begin{CJK}{UTF8}{gkai}难过、悲伤、伤心、悲痛、痛心\end{CJK} \\ 
 & Sad, sorrowful, grieved, distressed, pained \\
 Anger  & \begin{CJK}{UTF8}{gkai}愤怒、生气、气愤、恼火、恼怒\end{CJK} \\
 & Angry, annoy, indignant, furious, irritated \\
  Fear   & \begin{CJK}{UTF8}{gkai}害怕、恐惧、恐慌、畏惧、提心吊胆\end{CJK} \\
 & Fear, afraid, scare,  dread, frightened \\
  Disgust   & \begin{CJK}{UTF8}{gkai}讨厌、仇恨、厌恶、痛恨、怨恨\end{CJK} \\
 & Disgust, hatred, detest, abhor, grudge \\
  Surprise    & \begin{CJK}{UTF8}{gkai}惊讶、震惊、大吃一惊、惊奇、难以置信\end{CJK} \\
 & surprised, shocked, astonished, amazed, unbelievable \\ \hline
\end{tabular}
\caption{ Emotion category and emotion keywords}
\end{table*}

\subsubsection{Data source. }The National Language Resources Dynamic Circulation Corpus (DCC) 2005-2015\footnote{https://dcc.blcu.edu.cn}. The news text is more formal and complete so it is more likely that causes and actions appear in the same news text.

\subsubsection{Extraction.} We extract the passages with emotion keywords from DCC. In addition to the sentences including emotion keywords, three preceding clauses and three following clauses are kept as the context. Not all the extracted passages meet our requirement at the first stage, so we remove sentences including that: 1) are non-emotional; 2) have no experiencer; 3) don’t have emotion causes nor emotion actions; 4) have two or more emotion keywords.

\subsection{Annotation scheme}
Annotation format is the W3C Emotion Markup Language (EML) format and we made a slightly change for our task. The basic XML tags are: 1)emotion cause is marked by $\langle {cause}\rangle$. 2)emotion keyword is marked by 
$\langle {keyword}\rangle$. 3)emotion action is marked by $\langle {action}\rangle$. 4)experiencer is marked by $\langle {experiencer}\rangle$.

One emotion may have more than one corresponding emotion causes and actions, so (cause) and (action) tags have a “id” attribute to mark the number of causes and actions. There are two types of cause: noun/noun phrase and verb/verb phrase, so (cause) tag have a “type” attribute to mark cause type. Figure 2 shows two examples in the corpus, presented by the original simplified Chinese, followed by its English translation.

\begin{figure}[t]
    \centering
    \includegraphics[width=3in]{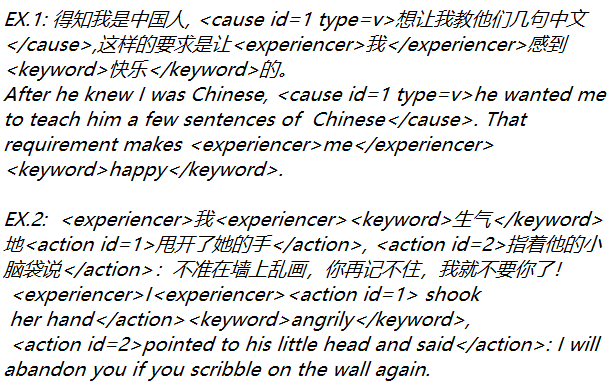}
    \caption{Examples of annotated sentences}
\end{figure}

\textbf{Annotation Procedure.} For each emotion keyword in each emotional sentence, two annotators manually annotate the cause(es), action(es) and experiencer independently. To each inconsistent sentence we involve a third annotator as the arbitrator. 

In order to balance the number of each emotion category and each emotion keyword, for each emotion category, we set the upper limit at about 1,700 instances, and for each emotion keyword, at about 300 instances. Finally we get 10,603 annotated sentences. Table 2 shows the sentences distribution of CEAC in each category.

\section{Statistics and Analysis}
\subsection{Data Distribution.}
\begin{table}[tbp]
    \centering
    \begin{tabular}{|c|c|} \hline
     Emotion category & Sentence number  \\ \hline
Happiness  & 1773 \\ \hline Fear  & 1748 \\ \hline
Sadness  & 1805 \\ \hline Disgust  & 1785 \\ \hline
Anger  & 1688 \\ \hline Surprise  & 1804 \\ \hline
    \end{tabular}
    \caption{The number of sentences in each category}
\end{table}

In CEAC, there are some sentences only containing causes, some sentences only containing actions, and sentences containing both. Table 3 shows the numbers of sentences of each type. It shows that about 77\% of sentences contain only causes and 80.0\% of clauses contain causes. There are very few sentences containing only actions, and very few clause containing causes and actions both. As show in the figure 3, we select an example for each type.

\begin{figure}[h]
    \centering
    \includegraphics[width=3in]{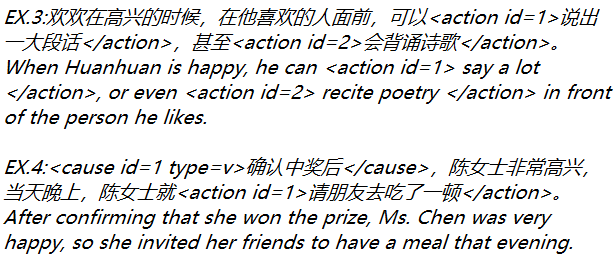}
    \caption{Clause containing only actions and containing causes and actions both}
\end{figure}

\begin{table}[h]
    \centering
    \resizebox{220pt}{12mm}{
    \begin{tabular}{|c|c|c|} \hline
    Item & Sentence & Clause \\ \hline
    just cause & 8167  (77.0\%) & 12782(80.0\%) \\
just action & 230    (2.2\%) &  3089  (19.3\%) \\
Cause \& action & 2206  (20.8\%) & 111    (7.0\%) \\
Total & 10603 & 15982 \\ \hline
    \end{tabular}}
    \caption{Distribution of sentence types}
\end{table}

Table 4\footnote{There are 111 clauses containing both cause and action, so the number of clauses is less than the number of events(16093)} shows the distribution of cause position and action position. Emotion causes appear much more in front of emotion keywords and Emotion actions appear vice verse. This is because news texts focus on narrative integrity and logic. In addition, time logical narration conforms to human's thinking habit, so it is a narrative way to describe the cause first and then the result in the text. 

\begin{table}[h]
    \centering
    \begin{tabular}{|c|c|c|} \hline
    Position  & Cause \% & Action \%  \\ \hline
    Previous 3 clauses & 556(4.3) & 32(1) \\
Previous 2 clauses & 1404(10.9) & 40(1.3) \\
Previous 1 clauses & 4554(35.3) & 96(3) \\
In the same clauses & 3897(30.2) & 782(24.4) \\
Next 1 clauses & 1172(9.1) & 1335(41.7) \\ 
Next 2 clauses & 572(4.5) & 521(16.3) \\
Next 3 clauses & 300(2.3) & 229(7.2)\\
Other & 438(3.4) & 165(5.1) \\
Total & 12893 & 3200 \\ \hline
    \end{tabular}
    \caption{Distribution of cause position and action position}
\end{table}

In emotion cause type distribution, verbal causes account for 75.2\%, as shown in table 5. In addition, we found all the emotion actions are verb/verb phrases.

\begin{table}[h]
    \centering
    \begin{tabular}{|c|c|c|} \hline
    Cause type & Number  & Percent  \\ \hline
    Noun/noun phrase & 3202 & 24.8\% \\
Verb/verb phrase & 9691 & 75.2\% \\ \hline
    \end{tabular}
    \caption{Distribution of cause type}
\end{table}

\textbf{Agreement.} In order to get high quality annotated examples, we trained the annotators strictly before annotating and allowed them to discard the difficult sentences. We use the same inter-annotator agreement method of \cite{gui2016event}. We reached 0.8201 for the Kappa value at clause level which is lower than \cite{gui2016event} because we need to label emotion action besides emotion cause.

\textbf{Inconsistent analysis.} We also analyzed the inconsistent sentences and find that the following situations may lead to inconsistent results. In the examples, the wrong annotations are marked with “*" and the correct annotations are marked with “\#".

1) Incorrectly annotate the condition of cause as emotion cause.

EX.5:(\begin{CJK}{UTF8}{gkai}郭平原言自家*受皇帝旌表*，\# 不能报答 \#，因而悲伤。\end{CJK} 

\textit{Guo Pingyuan said that *he was blessed by the emperor*, but \# he could not repay \#, so he was sad.})

\noindent “he could not repay” is the reason why he feel sad. Though “he was blessed by the emperor” is the premise of  sadness, there is no causal relationship, so we don’t think it is emotion cause.

2) The annotator incorrectly annotates actions that are contrary to emotions as emotion actions.

EX.8: (\begin{CJK}{UTF8}{gkai}这些痛失亲人的战友们，依然忍着悲伤，*继续战斗在抗震救灾的第一线*。\end{CJK} 

\textit{These comrades who lost their loved ones still endured sorrow and *fighted against earthquake* .})

\noindent “fighted against earthquake” isn’t the action caused by sorrow but caused by repressing sorrow, so it is contrary to sorrow.

3) Actions that occur with emotions may be mistakenly annotated as emotion action.

EX.9:( \begin{CJK}{UTF8}{gkai}一般民众在痛恨不良商家非法添加的同时，*越来越关注其国家标准允许的各种添加剂所带来的可能的危害了*。\end{CJK} 

\textit{While the general public hates the illegal addition of bad businesses, they *are paying more and more attention to the possible harm caused by various additives allowed by their national standards*.})

\noindent Because of the simultaneous occurrence of “hate” and “are paying more and more attention to the possible”, so the latter is not annotated as emotion action.

\section{Task}

\subsection{Task Definition}
\textbf{Cause-Emotion-Action Relation Extraction (Emotion Causality)}. We define the CEA relation extraction task as extracting or filling the slots in the triple (Cause, Emotion, Action) in a given text as below:

Given a text $W=\{w_{1},w_{2},...,w_{i},...,w_{n}\} $, where $w_{i}(i=1,..,n)$ is the word in the text, a triple $(C,E,A)$ need to be extracted as CEA relation, where $C=\{w_{1},...w_{i},...,w_{nc} \mid w_{i} \in W, 0 \leq nc<n\}, A=\{w_{1},...w_{j},...,w_{na} \mid w_{j} \in W, 0 \leq na<n \}$, are continuous word sequence in text $W$ respectively, $E \in EKSet$ is a emotion word. To our paper, $ESet=\{Happy, Sad,..., unbelievable \}$ is the emotion keywords listed in table 1.

\textbf{Emotion Inference}. We define the emotion Inference task as predict an emotion category by given an event of action/cause as below:

Given a Cause and Action event tuple (C, A), where  $C=\{w_{1},...w_{i},...,w_{nc} \mid w_{i} \in W, nc\geq 0\}, A=\{w_{1},...w_{j},...,w_{na} \mid w_{j} \in W, na\geq 0 \}$, are continuous word sequence respectively, and a given emotion category set $E = \{ Anger, Disgust, Fear, Happiness, Sadness, \\ Surprise \}$ , it needs to select a emotion category $e \in E$ as inference answer, i.e. $ \left \{ C,A \right \} \rightarrow  e$, ($e \in E$).


\subsection{Baseline Model}
\textbf{Cause-Emotion-Action Relation Extraction}. We regard this task as a sequence labeling problem and use Bi-LSTM + CRF model\cite{Huang2015Bidirectional}. It uses bidirectional LSTM to encode and add CRF layer on the top model structure. Based on the  conditional model , CRF model can mark the new observation sequence x by selecting the label sequence y that maximizes the conditional probability $P_{\left ( y|x \right)}$.

\textbf{Emotion Inference}. We treat it as a typical classification task. We use LSTM model to encode cause events and action events respectively. When using both cause events and action events as model input, we splice different LSTM’s (one for cause events and another one for action events) final hidden states into one vector. Then feed this vector to an softmax layer and predict the final result.

\section{Experiment}
\subsection{Dataset \& Hyperparameters}

\textbf{Cause-Emotion-Action Relation Extraction}. We directly use the 10603 texts in the CEAC data set as experimental data. The training set and test set are divided in a 4:1 ratio.

For LSTM+CRF model, char embeddings are random generate from [-0.25, 0.25] and they will be fixed during training. The sizes of these char embeddings are 300. The hidden size of three LSTMs are set to 300. The max epoch and size of batch are set to 40, 64, respectly. The Adam has been used to update parameters with learning rate 0.001 in our experiments.

\textbf{Emotion Inference}. For this task,we use the 10603 texts in the CEAC dataset as experimental data. The training set and test set are divided in a 4:1 ratio.

For LSTM model, word embeddings are pre-train on Wikipedia with embedding size 200. And they will not be update during training. The hidden size of three LSTMs are set to 200. The max epoch and size of batch are set to 16, 200, respectly. The Adam has been used to update parameters with learning rate 0.0005 in our experiments.

\subsection{Result} 
\textbf{Cause-Emotion-Action Relation Extraction}. As shown in the table 6, the result of extracting CEA relation is poor, which reflects the difficulty of the task to some extent. In addition to the performance of the test set, We also use 2206 sentences containing both causes and actions to see whether given actions can improve the extraction result of the causes and whether given causes can improve the extraction result of the actions. The training set and test set are divided in a 4:1 ratio. As we can see in the table 7, “cause \& action"item is the result of the detection of cause/action given corresponding cause/action. Obviously, after adding actions, there is 0.01 improvement in F-measure at the experiment of cause detection. After adding cause, there is 0.03 improvement in F-measure at the experiment of action detection. In a word, the result of detection will be improved by adding the corresponding action or cause information, especially in \textit{surprise}, which increased 0.16 in action detection after adding the cause.

\begin{table}[t]
    \centering
    \resizebox{220pt}{26mm}{
    \begin{tabular}{cccc|ccc} \hline
    \multicolumn{1}{c}{\textbf{ }} &
    \multicolumn{6}{c}{\textbf{Test Data}} \\ \hline
    \multicolumn{1}{c}{\textbf{}} & 
    \multicolumn{3}{c|}{\textbf{Cause}} & 
    \multicolumn{3}{c}{\textbf{Action}} \\ \hline
    & Pre. & Rec. & F. & Pre. & Rec. & F.\\ \hline
    Majority & 0.03 & 1.0 & 0.06  & 0.15 & 1.0 & 0.26  \\
    ALL & 0.55 & 0.53 & 0.54 & 0.48 & 0.44 & 0.46 \\ 
    Anger & 0.50 & 0.48 & 0.49 & 0.57 & 0.53 & 0.55  \\
    Disgust & 0.55 & 0.49 & 0.52 & 0.46 & 0.39 & 0.42  \\
    Fear & 0.52 & 0.50 & 0.51 & 0.36 & 0.33 & 0.34 \\
    Happiness & 0.52 & 0.49 & 0.50 & 0.53 & 0.45 & 0.49 \\
    Sadness & 0.61 & 0.59 & 0.60 & 0.46 & 0.44 & 0.45  \\
    Surprise & 0.61 & 0.59 & 0.60 & 0.37& 0.36 & 0.36  \\ \hline
    \end{tabular}}
    \caption{Result of the Cause-Emotion-Action Relation Extraction task}
\end{table}	

\begin{table*}[]
\centering
\resizebox{450pt}{25mm}{
\begin{tabular}{ccccccc|cccccc}
\hline
\multicolumn{7}{c|}{\textbf{CAUSE \& ACTION}}                                                                        & \multicolumn{6}{c}{\textbf{CAUSE \& ACTION}}                                         \\ \hline
\multicolumn{4}{c}{\textbf{cause}}                                                   & \multicolumn{3}{c|}{\textbf{action}} & \multicolumn{3}{c}{\textbf{cause}}                   & \multicolumn{3}{c}{\textbf{action}} \\ \hline
\multicolumn{1}{c|}{}         & Precision   & Recall & \multicolumn{1}{c|}{F1}     & Precision     & Recall   & F1      & Precision   & Recall & \multicolumn{1}{c|}{F1}     & Precision     & recall  & F1      \\ \hline
\multicolumn{1}{c|}{surprise} & 31.25 & 31.25  & \multicolumn{1}{c|}{31.25} & 61.9    & 50.0       & 55.32  & 28.57 & 25.0     & \multicolumn{1}{c|}{26.67} & 45.0      & 34.62   & 39.13  \\ \cline{2-13} 
\multicolumn{1}{c|}{disgust}  & 38.71 & 37.89  & \multicolumn{1}{c|}{38.3}  & 47.93   & 45.31    & 46.59  & 38.32 & 34.74  & \multicolumn{1}{c|}{36.67} & 44.63   & 42.19   & 43.37  \\ \cline{2-13} 
\multicolumn{1}{c|}{fear}     & 40.0    & 38.78  & \multicolumn{1}{c|}{39.38} & 42.11   & 36.36    & 39.02  & 44.3  & 35.71  & \multicolumn{1}{c|}{39.55} & 43.48   & 36.36   & 39.6   \\ \cline{2-13} 
\multicolumn{1}{c|}{happy}    & 46.15 & 42.86  & \multicolumn{1}{c|}{44.44} & 44.68   & 41.18    & 42.86  & 48.65 & 42.86  & \multicolumn{1}{c|}{45.57} & 46.81   & 43.14   & 44.9   \\ \cline{2-13} 
\multicolumn{1}{c|}{sadness}  & 48.05 & 50.68  & \multicolumn{1}{c|}{49.33} & 66.22   & 58.33    & 62.03  & 56.06 & 50.68  & \multicolumn{1}{c|}{53.24} & 53.75   & 51.19   & 52.44  \\ \cline{2-13} 
\multicolumn{1}{c|}{anger}    & 49.36 & 50.0     & \multicolumn{1}{c|}{49.68} & 62.8    & 58.19    & 60.41  & 43.68 & 45.45  & \multicolumn{1}{c|}{44.44} & 58.86   & 58.19   & 58.52  \\ \cline{2-13} 
\multicolumn{1}{c|}{all}      & 44.33 & 44.14  & \multicolumn{1}{c|}{44.23} & 54.41   & 49.31    & 51.73  & 44.57 & 41.21  & \multicolumn{1}{c|}{42.83} & 50.65   & 47.05   & 48.78  \\ \hline
\end{tabular}}
\caption{Result of the Cause-Emotion-Action Relation Extraction task on the
dataset which both contain cause and action}
\end{table*}

\textbf{Emotion Inference}. We also extract part of results from the dataset that contained both causes and actions for comparative analysis. From the table 8, we can see that the differences between the categories of emotions in task 2 are not as great as in task 1. However, the model with more information (both cause and action) can still achieve better results (5\% higher than overall test data set). The result of inferring three categories (Anger, Fear and Sadness ) has been significantly enhanced.

\section{Discussion}
In this section, we will analyze the experimental results and make some discussion on it.

As explained in the introduction section, the triple (cause, emotion, action) have multiple causal chains: 1) Emotions and cause events are causal. 2) Emotions lead to action events. 3) Causes and actions are also causal. Based on this, we conducted two groups of experiments: task1 and task2. In Task 1, we find that the result of extracting emotion triples CEA is poor. This is because the task itself is hard and there are many extracting contents. We also find that the result of detection will be improved by adding the corresponding action or cause information, for example, \textit{“Wang Yan was angry when she found out that her husband had derailed, so she decided to divorce her husband"}. In this example, the model can detect \textit{decided to divorce her husband"} as action more easier after given the cause \textit{“she found out that her husband had derailed"}. In task 2, after adding more information, the results of the experiment in some emotions are improved, such as \textit{anger, fear, and sad}. However, there are also some exceptions, such as \textit{disgust, happiness, surprise}. After analyzing the data, we found that the causes of anger, fear and sadness are similar. For example, the events \textit{“he hurt me"} can both lead to anger, fear and sadness. Therefore, the result of inferring emotions by causes alone is ineffective. When we put in the actions, for example, \textit{“retaliate"}, \textit{“dodge"} and \textit{“cry"}, we can distinguish these three emotions and the result of experiment is improved. 
The imbalance of data that contain both causes and actions affect the results of the experiment. As shown in the figure 4, we can see that when the proportion of data contains both causes and actions changes, the improvement also changes regularly. For example, in \textit{surprise}, the reason why the result of the whole model becomes even worse after adding the action is that the rate of data that contains actions is low, so the action becomes useless noise in that case.  

\begin{table*}[t]
    \centering
    \begin{tabular}{c|cccc|cccc} \hline
    \multicolumn{1}{c}{\textbf{ }} & \multicolumn{4}{c}{\textbf{ALL}} & 
    \multicolumn{4}{c}{\textbf{ACTION}} \\ \hline
	&	Precision & Recall & F1 & Support & Precision & Recall & F1 & Support\\ \hline

ALL & 0.55  & 0.55  & 0.55  & 2110 & 0.63  & 0.51  & 0.50  & 39\\
Anger & 0.54  & 0.45  & 0.49  & 345 & 0.71  & 0.28  & 0.40  & 18\\
Disgust & 0.55  & 0.48  & 0.51  & 357 & 0.42  & 0.73  & 0.53  & 11\\
Fear & 0.50  & 0.52  & 0.51  & 321 & 0.33  & 1.00  & 0.50  & 2\\
Happiness & 0.60  & 0.66  & 0.63  & 335 & 0.60  & 0.75  & 0.67  & 4\\
Sadness & 0.60  & 0.65  & 0.62  & 359 & 1.00  & 0.50  & 0.67  & 4\\
Surprise & 0.50  & 0.54  & 0.52  & 393 & NULL & NULL & NULL &0\\ \hline
\multicolumn{1}{c}{\textbf{ }} &
    \multicolumn{4}{c}{\textbf{CAUSE}} & 
    \multicolumn{4}{c}{\textbf{CAUSE \& ACTION}} \\ \hline
     & Precision & Recall & F1 & Support & Precision & Recall & F1 & Support \\ \hline

ALL & 0.53  & 0.53  & 0.53  & 1627 & 0.61  & 0.60  & 0.60  & 444 \\
Anger & 0.41  & 0.30  & 0.34  & 185 & 0.66  & 0.66  & 0.66  & 142 \\
Disgust & 0.57  & 0.47  & 0.52  & 268 & 0.51  & 0.47  & 0.49  & 78 \\
Fear & 0.45  & 0.50  & 0.47  & 228 & 0.68  & 0.57  & 0.62  & 91 \\
Happiness & 0.60  & 0.68  & 0.64  & 291 & 0.62  & 0.50  & 0.56  & 40 \\
Sadness & 0.60  & 0.60  & 0.60  & 286 & 0.59  & 0.83  & 0.69  & 69 \\
Surprise & 0.51  & 0.55 & 0.53  & 369 & 0.32  & 0.33  & 0.33  & 23 \\ \hline
    \end{tabular}
    \caption{Results of emotion inference task}
\end{table*}

\begin{figure}[h]
    \centering
    \includegraphics[width=3in]{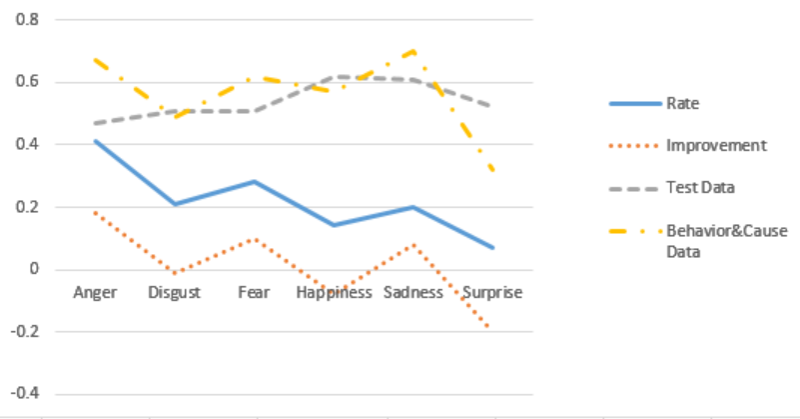}
    \caption{Visualization of the performance changes along with the ratio of data which contains both causes and actions}
\end{figure}

\section{Related Works}
We only list emotion event/cause data source-related works here. 

\cite{tokuhisa2008emotion} first defined emotion-provoking event and constructed a corpus in Japanese using massive examples extracted from the web, then did sentiment polarity classification and emotion classification. \cite{vu2014acquiring}worked on creating prevalence-ranked dictionaries of emotion-provoking events through both manual labor and automatic information extraction. 

\cite{lee2010emotion} first proposed a task on emotion cause detection. They manually constructed a corpus from Academia Sinica Balanced Chinese Corpus. \cite{gui2016event} built a dataset using SINA city news then propose an event-driven emotion cause extraction method using multi-kernel SVMs. \cite{ghazi2015detecting} directly selected the emotions-directed frames in FrameNet to build an English emotion cause (or stimulus) corpus then used CRFs to detect emotion causes. Some study \cite{gui2014emotion} designed corpus through annotating the emotion cause expressions in Chinese Weibo and extended the rule based method to informal text in Weibo text. \cite{xiyaochengargument} focused on current/original-subtweet-based emotion detection and annotated a multiple-user structure. \cite{gao2017overview} organized NTCIR-13 ECA (emotion cause analysis) task. It designed two subtasks including emotion cause detection subtask and emotion cause extraction subtask.

\cite{deng2013benefactive} presented an annotation scheme for events that negatively or positively affect entities (benefactive/malefactive events). Then \cite{choi2014lexical} constructed two sense-level lexicon of benefactive and malefactive events for opinion inference. 

\cite{ding2016acquiring} defined affective events as events that are typically associated with a positive or negative emotional state and aim to automatically acquire knowledge of stereotypically positive and negative events from personal blogs. \cite{ding2018weakly} defined a set of categories human need to explain the affect of events. They also manually added manual annotations of human need categories to a previous collection of affective events.

\cite{rashkin2018event2mind} proposed Event2Mind to supporting commonsense inference on events with a specific focus on modeling stereotypical intents and reactions of people, described in short free-form text.

\section{Conclusion}
In this paper, first we define emotion action and put it into emotion causality so that cause, emotion and action comprise an integral cause-effect chain. Then we define and investigate emotion causality and emotion inference tasks. We manually label a large-scale corpus CEAC to support the two tasks. Finally, we report baseline performance on the tasks and it shows that: for the emotion causality, the performance of the state-of-the-art sequence labeling model are still too difficult to achieve good performance; for the emotion inference, the popular neuron model can compose embedding representations of previously unseen events and possible emotion causes and for both tasks, the emotion action does affect the result of experiment. 

There is still much room for improvement in emotion causality and emotion inference task. In addition, we cannot make soundly analysis on all the experimental results now because the imbalance distribution of emotion cause and emotion action, so we aim to release 50,000 instances in the future which we believe it can significantly boost the study in both emotion causality and emotion inference research area. We are currently releasing 10,603 samples with 16,093 events to inspire work in emotion causality and emotion inference or other related task and along with gathering feedback from the research community.
\bibliography{naaclhlt2019}
\bibliographystyle{acl_natbib}

\end{document}